\documentclass[conference]{IEEEtran}

\usepackage{cite}
\usepackage{amsmath,amssymb}
\usepackage{graphicx}
\usepackage{booktabs}
\usepackage{xcolor}
\usepackage{url}
\usepackage{tikz}
\usetikzlibrary{arrows.meta,positioning,fit,calc}

\begin{document}

\title{Backbone Matters: Benchmarking Encoder--Decoder Networks for Semantic Segmentation of Driving Scenes on BDD100K}

\author{%
\IEEEauthorblockN{Ehsan Rassekh}
\IEEEauthorblockA{Institute for Advanced Studies in Basic Sciences (IASBS), Zanjan, Iran\\
Email: rassekh.ehsan@gmail.com, rassekh.ehsan@spes.uniud.it}
}

\maketitle

\begin{abstract}
Semantic segmentation gives an autonomous vehicle a dense, pixel-level description of its surroundings and is a core part of scene understanding. Most segmentation networks consist of an encoder (the backbone) and a decoder, yet the backbone is often chosen without a systematic comparison. This work studies this choice on the BDD100K driving dataset. Four widely used decoders (U-Net, FPN, LinkNet and PSPNet) are combined with 13 backbones from six families (ResNet, VGG, DenseNet, Inception, MobileNet and EfficientNet), and all 52 models are trained with one common protocol. The best model, an FPN decoder with an EfficientNet-B4 backbone, reaches 62.6\% mean intersection over union (mIoU), 89.8\% accuracy and a loss of 0.092. Only two of the 52 models, both FPN with EfficientNet, exceed the 56.9\% mIoU reported with the BDD100K benchmark. The effect of the backbone depends strongly on the decoder: the same EfficientNet-B4 gives 42.6\% mIoU with PSPNet but 62.6\% with FPN, and the gap between the best and the worst backbone ranges from 12.0 points (U-Net) to 42.5 points (FPN). The results and the scope of this study are discussed.
\end{abstract}

\begin{IEEEkeywords}
semantic segmentation, scene understanding, autonomous driving, backbone, BDD100K, feature pyramid network, EfficientNet
\end{IEEEkeywords}

\section{Introduction}
Road safety is one of the strongest motivations for self-driving technology. Traffic accidents cause about 1.3 million deaths and 20--50 million injuries every year \cite{asirt}, and more than 90\% of crashes are linked to human error \cite{singh2015critical}. A self-driving system is usually described as a pipeline of sensing, perception and decision making \cite{badue2021self}. Perception is a part where deep learning can help a lot: a vehicle can only take safe decisions if it understands the road, the lane markings, the traffic signs, and the vehicles and pedestrians around it.

\emph{Semantic segmentation} supports this understanding by assigning a class label (for example ``road'', ``sky'', ``car'' or ``pedestrian'') to every pixel of a camera image \cite{guo2021survey,garcia2017review}. For driving, the data must cover many lighting and weather conditions, and the model must find object boundaries precisely. Modern segmentation networks follow an encoder--decoder design. The encoder, called the \emph{backbone}, extracts features from the image; the decoder turns these features back into a full-resolution label map. Many different backbones exist, but in practice one is often chosen in advance and the comparison focuses on the decoder.

This paper asks a simple question: \emph{how much does the choice of backbone change the quality of semantic segmentation for driving scenes, and does the answer depend on the decoder?} To answer it, this work presents a controlled study on BDD100K \cite{yu2020bdd100k}, building on earlier work \cite{rassekh2025semantic}. The contributions of this work are:
\begin{itemize}
  \item A systematic benchmark of 52 segmentation models: four decoders (U-Net, FPN, LinkNet, PSPNet) combined with 13 backbones from six families, all trained with the same protocol and reported with mIoU, pixel accuracy and loss.
  \item The finding that FPN with an EfficientNet backbone is clearly the best combination in this study (62.6\% mIoU with EfficientNet-B4), while the same backbone performs at the level of the other backbones with the other three decoders.
  \item An analysis of backbone sensitivity, of the relation between mIoU, accuracy and loss, and of the scope of the study, which points to concrete next steps.
\end{itemize}

\section{Related Work}
\subsection{Segmentation Architectures}
Fully convolutional networks (FCN) \cite{long2015fcn} showed that a classification network can be turned into a dense predictor. Encoder--decoder networks such as SegNet \cite{badrinarayanan2017segnet} and U-Net \cite{ronneberger2015unet} then added a decoder that recovers spatial detail; U-Net also copies encoder features to the decoder by skip connections and was first designed for biomedical images. Several works improve the use of context. PSPNet \cite{zhao2017pspnet} pools the feature map at several scales to collect global context. The DeepLab family uses atrous (dilated) convolution and atrous spatial pyramid pooling \cite{chen2018deeplab,chen2017rethinking}, and DeepLabv3+ adds a light decoder \cite{chen2018encoder}. HRNet \cite{wang2021hrnet} keeps a high-resolution representation through the whole network. Feature Pyramid Networks (FPN) \cite{lin2017fpn} build a pyramid of features with a top-down path and lateral connections; they were proposed for object detection and are also effective for segmentation \cite{kirillov2019panoptic}. LinkNet \cite{chaurasia2017linknet} was designed for efficiency: it links each encoder block to the matching decoder block to keep the number of parameters low. More recent work uses vision transformers and attention-based decoders, such as SegFormer \cite{xie2021segformer} and Mask2Former \cite{cheng2022mask2former}, and modernized convolutional backbones such as ConvNeXt \cite{liu2022convnet}. This paper focuses on classical convolutional encoder--decoder networks, which remain a common baseline. Surveys of deep segmentation are given in \cite{minaee2022image,garcia2017review}; for driving scenes, see \cite{guo2021survey,feng2021deep}, and \cite{siam2018comparative} for a comparison of real-time networks.

\subsection{Backbones}
The encoder of a segmentation network is normally an image-classification network used as a feature extractor. ResNet \cite{he2016resnet} uses residual connections to train deep networks, VGG \cite{simonyan2014vgg} stacks small convolution filters, DenseNet \cite{huang2017densenet} connects every layer to all later layers in a block, and Inception \cite{szegedy2016inception,szegedy2017inceptionresnet} uses parallel filters of different sizes. MobileNet \cite{howard2017mobilenets,sandler2018mobilenetv2} uses depthwise separable convolutions to reduce computation. EfficientNet \cite{tan2019efficientnet} scales network depth, width and input resolution together with one compound coefficient instead of scaling a single dimension. Using pretrained backbones (transfer learning) is standard practice, because it lets the segmentation model reuse features learned on a large classification dataset such as ImageNet \cite{deng2009imagenet,yosinski2014transferable}. Better classification backbones usually transfer better to other tasks, but the relation is not perfect \cite{kornblith2019better}, which is one more reason to compare backbones directly.

\subsection{Driving Datasets}
Public datasets for urban scenes include CamVid \cite{brostow2009camvid}, KITTI \cite{geiger2012kitti}, Cityscapes \cite{cordts2016cityscapes} and Mapillary Vistas \cite{neuhold2017mapillary}. BDD100K \cite{yu2020bdd100k} contains 100K driving videos with ten tasks, recorded in different cities, at different times of day and in different weather. The BDD100K paper reports 56.9\% mIoU for a semantic segmentation model trained on segmentation data alone, and 58.3\% when it is trained jointly with object detection data. The first number is used as a reference point in Section~\ref{sec:results}.

\section{Method}
\subsection{Overall Design}
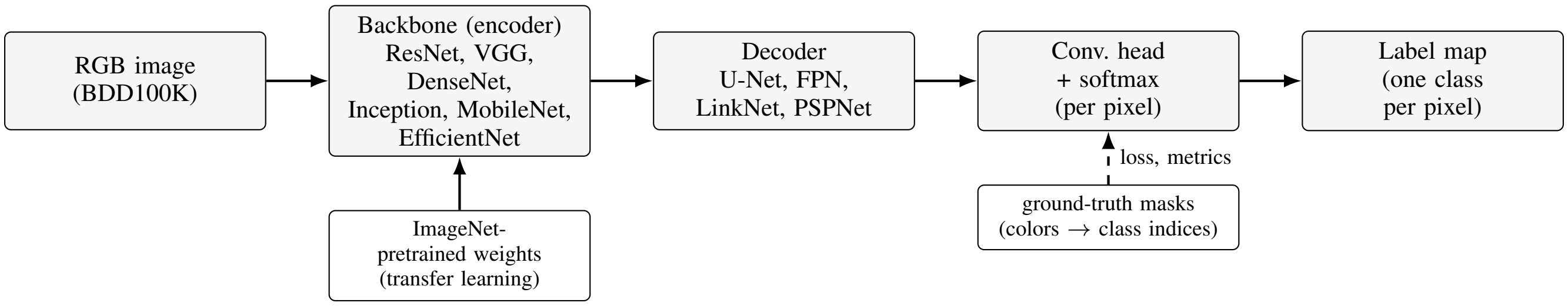
\begin{figure*}[t]
\centering
\begin{tikzpicture}[
  node distance=7mm,
  blk/.style={draw, rounded corners=2pt, align=center, minimum height=11mm, text width=2.7cm, font=\footnotesize, fill=black!4},
  small/.style={draw, rounded corners=2pt, align=center, minimum height=7mm, text width=2.7cm, font=\scriptsize, fill=white},
  arr/.style={-{Latex[length=2mm]}, thick}
]
\node[blk] (img) {RGB image\\(BDD100K)};
\node[blk, right=of img] (bb) {Backbone (encoder)\\ResNet, VGG, DenseNet,\\Inception, MobileNet,\\EfficientNet};
\node[blk, right=of bb] (dec) {Decoder\\U-Net, FPN,\\LinkNet, PSPNet};
\node[blk, right=of dec] (head) {Conv.\ head\\+ softmax\\(per pixel)};
\node[blk, right=of head] (out) {Label map\\(one class\\per pixel)};
\draw[arr] (img) -- (bb);
\draw[arr] (bb) -- (dec);
\draw[arr] (dec) -- (head);
\draw[arr] (head) -- (out);
\node[small, below=6mm of bb] (tl) {ImageNet-pretrained weights\\(transfer learning)};
\draw[arr] (tl) -- (bb);
\node[small, below=6mm of head] (gt) {ground-truth masks\\(colors $\rightarrow$ class indices)};
\draw[arr, dashed] (gt.north) -- node[right, font=\scriptsize]{loss, metrics} (head.south);
\end{tikzpicture}
\caption{Pipeline used in this study. The backbone and the decoder are interchangeable; 13 backbones and four decoders give 52 models.}
\label{fig:pipeline}
\end{figure*}

Fig.~\ref{fig:pipeline} shows the pipeline. An RGB image is passed through a backbone that is initialized with pretrained weights. A decoder uses the backbone features to build a high-resolution representation, and a final convolution with a softmax activation gives one probability per class for each pixel. Any backbone can be combined with any decoder, so the pipeline allows a controlled comparison where only one component changes at a time.

\subsection{Decoders}
Four decoders are evaluated; they differ mainly in how they bring information from the encoder to the output.

\emph{U-Net} \cite{ronneberger2015unet} has a contracting path and a symmetric expanding path. At each step of the expanding path the feature map is upsampled with an up-convolution and concatenated with the feature map of the same resolution from the contracting path.

\begin{figure}[t]
\centering
\begin{tikzpicture}[font=\scriptsize,
  st/.style={draw, rounded corners=1pt, minimum height=5.5mm, fill=black!6},
  arr/.style={-{Latex[length=1.6mm]}, thick}]
\foreach \i/\y/\w in {2/0/2.0, 3/1.0/1.7, 4/2.0/1.4, 5/3.0/1.1}{
  \node[st, minimum width=\w cm] (C\i) at (0.9,\y) {$C_\i$};
  \node[st, minimum width=\w cm, fill=white] (P\i) at (4.1,\y) {$P_\i$};
}
\foreach \a/\b in {2/3, 3/4, 4/5}{
  \draw[arr] (C\a.north) -- (C\b.south);
  \draw[arr] (P\b.south) -- (P\a.north);
}
\foreach \i in {2,3,4,5}{
  \draw[arr] (C\i.east) -- node[above, pos=0.5]{$1{\times}1$} (P\i.west);
}
\node[draw, rounded corners=1pt, align=center, minimum height=3.9cm, text width=1.35cm] (head) at (6.9,1.5) {merge,\\conv.\ head,\\softmax};
\foreach \i in {2,3,4,5}{
  \draw[arr] (P\i.east) -- (head.west |- P\i.east);
}
\node at (0.9,3.65) {backbone (bottom-up)};
\node at (4.1,3.65) {top-down pathway};
\end{tikzpicture}
\caption{FPN decoder on top of a backbone. Each backbone stage $C_i$ is reduced with a $1{\times}1$ convolution and added to the $2{\times}$ upsampled map from the level above (the top level has no map above), which gives the pyramid maps $P_i$; these are merged for the final prediction.}
\label{fig:fpn}
\end{figure}
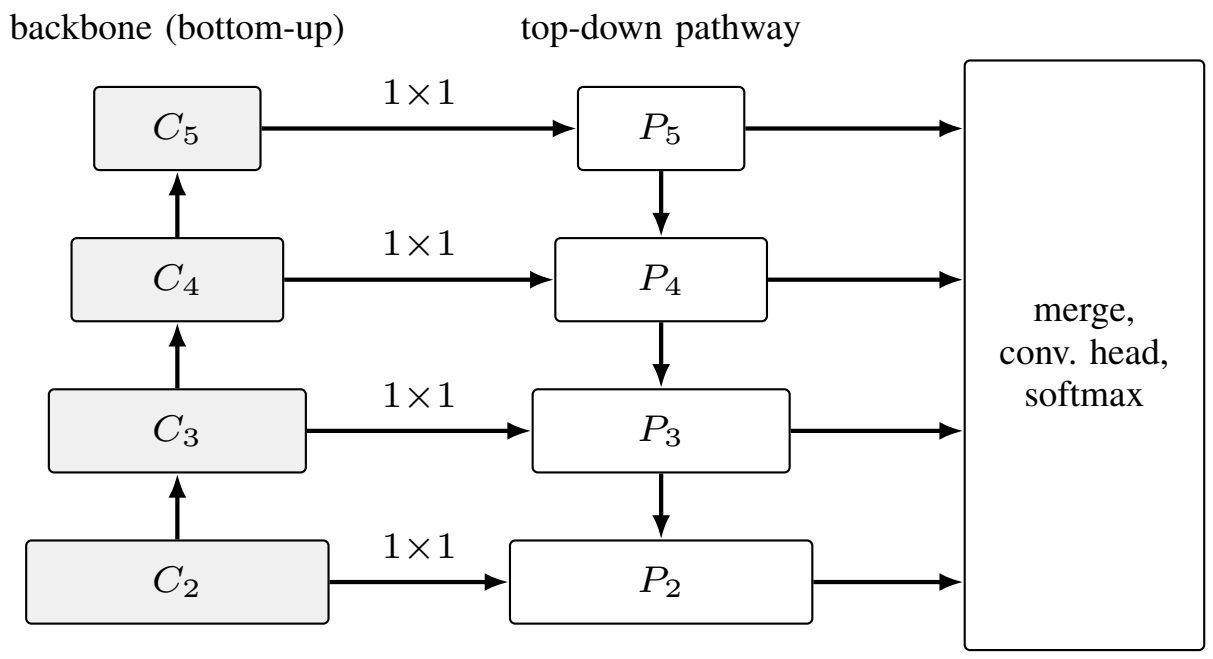

\emph{FPN} \cite{lin2017fpn} has a bottom-up pathway (the backbone), a top-down pathway and lateral connections (Fig.~\ref{fig:fpn}). The top-down pathway upsamples semantically strong but coarse maps by a factor of two with nearest-neighbor interpolation. Each lateral connection applies a $1{\times}1$ convolution to the backbone map of the same size and merges it with the upsampled map by element-wise addition.

\emph{LinkNet} \cite{chaurasia2017linknet} uses an initial $7{\times}7$ convolution with stride two, followed by residual encoder blocks. The input of every encoder block is linked to the output of the matching decoder block, which recovers spatial detail with few extra parameters.

\emph{PSPNet} \cite{zhao2017pspnet} applies a pyramid pooling module to the last feature map. The map is average-pooled into $1{\times}1$, $2{\times}2$, $3{\times}3$ and $6{\times}6$ sub-regions; each level is reduced with a $1{\times}1$ convolution, upsampled bilinearly, and concatenated with the original map to combine local and global context.

\subsection{Backbones and Transfer Learning}
The study uses 13 backbones: ResNet-34 and ResNet-50 \cite{he2016resnet}; VGG-16 and VGG-19 \cite{simonyan2014vgg}; DenseNet-121, -169 and -201 \cite{huang2017densenet}; Inception-v3 \cite{szegedy2016inception} and Inception-ResNet-v2 \cite{szegedy2017inceptionresnet}; MobileNet \cite{howard2017mobilenets} and MobileNetV2 \cite{sandler2018mobilenetv2}; and EfficientNet-B3 and -B4 \cite{tan2019efficientnet}. Every backbone is initialized with ImageNet-pretrained weights \cite{deng2009imagenet} and then trained for segmentation.

\subsection{Data Preparation}
The ground-truth masks of BDD100K are color images in which each class has a fixed RGB color. For training, each mask is converted into an integer label map by looking up the color of every pixel in a table of class colors. The table has 22 label indices, including the ``unlabeled'' and ``static'' indices, and every pixel receives the index of its matching color.

\subsection{Loss and Metrics}
Training minimizes the pixel-wise categorical cross-entropy,
\begin{equation}
\mathcal{L} = -\frac{1}{N}\sum_{p=1}^{N}\sum_{c=1}^{C} y_{p,c}\,\log \hat{y}_{p,c},
\end{equation}
where $N$ is the number of pixels, $C$ the number of classes, $y_{p,c}$ the one-hot ground truth and $\hat{y}_{p,c}$ the softmax output. Three quantities are reported. The first is the intersection over union (IoU) of class $c$ and its mean over classes,
\begin{equation}
\mathrm{IoU}_c = \frac{TP_c}{TP_c + FP_c + FN_c},\qquad
\mathrm{mIoU} = \frac{1}{C}\sum_{c=1}^{C}\mathrm{IoU}_c,
\end{equation}
where $TP_c$, $FP_c$ and $FN_c$ are the true positives, false positives and false negatives of class $c$. The second is the pixel accuracy, the fraction of correctly classified pixels. The third is the loss $\mathcal{L}$. Pixel accuracy alone is not enough to compare models, because frequent classes such as road and sky dominate it. mIoU gives every class the same weight and is therefore the main metric of this work.

\section{Experimental Setup}
\subsection{Data}
The study uses BDD100K \cite{yu2020bdd100k}, collected by the University of California, Berkeley. It contains 100K driving videos of about 40 seconds each, recorded at 720p and 30 fps with GPS/IMU data, on more than 50K rides in New York, the San Francisco Bay Area and other regions. It covers city streets, residential areas and highways, in different weather and at different times of day. Models are trained and evaluated on the labeled segmentation images, which are divided into 70\% training, 10\% validation and 20\% test images. The small CamVid dataset \cite{brostow2009camvid} (701 annotated frames) was also used as an auxiliary set; its size made it easy to test the models and to find implementation errors quickly. All results in this paper are on BDD100K. All reported results (mIoU, accuracy and loss) are computed on the test images.

\subsection{Implementation and Training}
All networks are implemented in Python with TensorFlow \cite{abadi2016tensorflow} and Keras, using an open-source Keras/TensorFlow library of segmentation networks. The training protocol follows \cite{rassekh2025semantic}, and all 52 models share the hyperparameters in Table~\ref{tab:hyper}. Training uses the Adam optimizer \cite{kingma2014adam}. The experiments ran on a server with an Intel Xeon Silver 4210R CPU (2.40\,GHz), 128\,GiB of memory, four Tesla K80 GPUs and one Tesla V100S GPU (32\,GB). Training and evaluating all 52 models took about 130 hours in total on this machine.

\begin{table}[t]
\caption{Training hyperparameters used for all models.}
\label{tab:hyper}
\centering
\small
\begin{tabular}{@{}ll@{}}
\toprule
Hyperparameter & Value \\
\midrule
Output activation & softmax \\
Loss & categorical cross-entropy \\
Optimizer & Adam \\
Learning rate & $1\times10^{-4}$ \\
Batch size & 8 \\
Epochs & at least 50 \\
IoU score / F-score threshold & 0.5 / 0.5 \\
\bottomrule
\end{tabular}
\end{table}

\section{Results and Discussion}
\label{sec:results}
\subsection{Main Results}
\begin{figure*}[t]
\centering
\includegraphics[width=\textwidth]{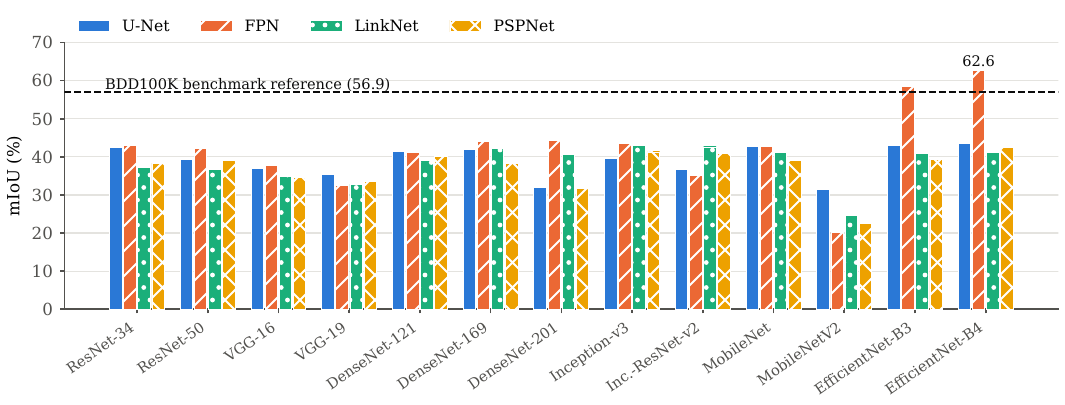}
\caption{mIoU (\%) for each backbone and decoder. The dashed line is the 56.9\% mIoU reported with the BDD100K benchmark for segmentation-only training \cite{yu2020bdd100k}. Only FPN with EfficientNet-B3 and -B4 is above this line.}
\label{fig:miou}
\end{figure*}

Table~\ref{tab:miou} and Fig.~\ref{fig:miou} give the mIoU of all 52 models, and Table~\ref{tab:accloss} gives their accuracy and loss. The best model is FPN with an EfficientNet-B4 backbone. It reaches 62.6\% mIoU and the highest accuracy (89.8\%), with a loss of 0.092. FPN with EfficientNet-B3 is second (58.6\% mIoU, 87.0\% accuracy). The next best model, FPN with DenseNet-201, reaches only 44.3\% mIoU, so the two EfficientNet models are separated from the rest of the field by more than 14 points.

\begin{table}[t]
\caption{mIoU (\%) of every decoder--backbone combination on BDD100K. The best backbone for each decoder is in bold.}
\label{tab:miou}
\centering
\small
\begin{tabular}{@{}lcccc@{}}
\toprule
Backbone & U-Net & FPN & LinkNet & PSPNet \\
\midrule
ResNet-34 & 42.6 & 42.9 & 37.3 & 38.3 \\
ResNet-50 & 39.4 & 42.3 & 36.7 & 38.9 \\
VGG-16 & 36.8 & 37.8 & 34.8 & 34.5 \\
VGG-19 & 35.3 & 32.5 & 32.6 & 33.6 \\
DenseNet-121 & 41.4 & 41.1 & 39.2 & 40.1 \\
DenseNet-169 & 41.9 & 44.0 & 42.3 & 38.3 \\
DenseNet-201 & 31.9 & 44.3 & 40.5 & 31.6 \\
Inception-v3 & 39.7 & 43.5 & \textbf{43.1} & 41.3 \\
Inc.-ResNet-v2 & 36.8 & 35.1 & 42.6 & 40.9 \\
MobileNet & 42.6 & 42.7 & 41.2 & 39.0 \\
MobileNetV2 & 31.4 & 20.1 & 24.5 & 22.6 \\
EfficientNet-B3 & 42.9 & 58.6 & 40.9 & 39.2 \\
EfficientNet-B4 & \textbf{43.5} & \textbf{62.6} & 41.2 & \textbf{42.6} \\
\bottomrule
\end{tabular}
\end{table}

\begin{table*}[t]
\caption{Test accuracy (Acc., \%) and loss of every model. The best accuracy and the lowest loss for each decoder are in bold.}
\label{tab:accloss}
\centering
\small
\begin{tabular}{@{}lcccccccc@{}}
\toprule
 & \multicolumn{2}{c}{U-Net} & \multicolumn{2}{c}{FPN} & \multicolumn{2}{c}{LinkNet} & \multicolumn{2}{c}{PSPNet}\\
\cmidrule(lr){2-3}\cmidrule(lr){4-5}\cmidrule(lr){6-7}\cmidrule(lr){8-9}
Backbone & Acc. & Loss & Acc. & Loss & Acc. & Loss & Acc. & Loss \\
\midrule
ResNet-34 & 84.3 & 0.144 & 84.8 & \textbf{0.084} & 78.3 & \textbf{0.102} & 82.2 & 0.175 \\
ResNet-50 & 81.8 & 0.170 & 83.9 & 0.102 & 81.1 & 0.199 & 82.6 & 0.179 \\
VGG-16 & 79.2 & 0.254 & 81.0 & 0.221 & 78.8 & 0.485 & 79.0 & 0.351 \\
VGG-19 & 78.6 & 0.270 & 77.5 & 0.364 & 77.3 & 0.429 & 76.8 & 0.354 \\
DenseNet-121 & 84.3 & 0.193 & 82.4 & 0.138 & 81.1 & 0.318 & 83.3 & 0.179 \\
DenseNet-169 & 84.8 & 0.173 & 85.6 & 0.102 & 84.8 & 0.209 & 82.3 & 0.235 \\
DenseNet-201 & 72.2 & 0.218 & 85.3 & 0.099 & 83.8 & 0.214 & 71.0 & 0.424 \\
Inception-v3 & 83.8 & 0.173 & 85.0 & 0.146 & 83.9 & 0.200 & 82.6 & 0.172 \\
Inc.-ResNet-v2 & 78.3 & 0.452 & 78.0 & 0.508 & \textbf{85.0} & 0.166 & 83.2 & \textbf{0.154} \\
MobileNet & 84.6 & \textbf{0.134} & 84.8 & 0.103 & 84.3 & 0.159 & 80.8 & 0.204 \\
MobileNetV2 & 70.2 & 0.209 & 43.8 & 0.207 & 53.8 & 0.315 & 56.8 & 0.363 \\
EfficientNet-B3 & 84.9 & 0.200 & 87.0 & 0.112 & 82.8 & 0.187 & 82.4 & 0.217 \\
EfficientNet-B4 & \textbf{85.8} & 0.152 & \textbf{89.8} & 0.092 & 83.8 & 0.166 & \textbf{83.9} & 0.198 \\
\bottomrule
\end{tabular}
\end{table*}

\subsection{Effect of the Backbone}
The backbone matters, but how much depends on the decoder. Within one decoder, the difference between the best and the worst backbone is 12.0 points for U-Net, 18.6 for LinkNet, 20.0 for PSPNet and 42.5 for FPN. MobileNetV2 is the worst backbone for all four decoders (20.1--31.4\% mIoU), and for FPN its accuracy falls to 43.8\%. This behavior deserves a closer analysis, for example with a tuned training setup; in the following comparison these runs are treated as outliers.

Among the other backbones the picture is more regular. VGG-19 is below VGG-16 for all four decoders, and the two VGG networks are the weakest backbones after MobileNetV2. A deeper backbone is not always better: ResNet-50 is below ResNet-34 with U-Net, FPN and LinkNet, and DenseNet-201 is the weakest DenseNet with U-Net and PSPNet. The light MobileNet is competitive with U-Net, LinkNet and PSPNet: it is within four points of the best backbone for each of them, which makes it interesting when computation is limited; measuring its inference time and model size is a useful next step.

\subsection{Interaction Between Backbone and Decoder}
\begin{table}[t]
\caption{Mean mIoU (\%) per backbone family: over the three decoders U-Net, LinkNet and PSPNet, and with FPN.}
\label{tab:family}
\centering
\small
\begin{tabular}{@{}lccc@{}}
\toprule
Family & Backbones & U-Net/LinkNet/PSPNet & FPN \\
\midrule
ResNet & 2 & 38.9 & 42.6 \\
VGG & 2 & 34.6 & 35.2 \\
DenseNet & 3 & 38.6 & 43.1 \\
Inception & 2 & 40.7 & 39.3 \\
MobileNet & 2 & 33.6 & 31.4 \\
EfficientNet & 2 & 41.7 & 60.6 \\
\bottomrule
\end{tabular}
\end{table}
The most striking result is the interaction between EfficientNet and FPN. Table~\ref{tab:family} averages the mIoU over the backbones of each family. For the five other families, the FPN average differs from the average of the other decoders by at most 4.5 points (from $-2.2$ to $+4.5$). For EfficientNet the difference is $+18.9$ points (60.6\% against 41.7\%).
With U-Net, LinkNet and PSPNet, EfficientNet-B3 and -B4 give 39--44\% mIoU, similar to most other backbones. With FPN they give 58.6\% and 62.6\%. The decoder alone does not explain this, because FPN with the other backbones gives 20--44\%; the backbone alone does not explain it either. A plausible explanation is that FPN merges features from several stages with lateral connections, which suits the multi-scale features of a compound-scaled network such as EfficientNet. Ablation studies and repeated training runs can test this explanation in future work.

\subsection{Accuracy, Loss and mIoU}
Over the 52 models, mIoU and accuracy rank the models almost in the same way (Spearman correlation 0.97), and mIoU and loss are also correlated (Spearman $-0.77$). However, the two measures have very different ranges. If MobileNetV2 and the two FPN--EfficientNet models are left out, 40 of the remaining 46 models have 78--86\% accuracy, while all of them have only 31.6--44.3\% mIoU. This is typical for imbalanced data: frequent classes cover most pixels and make the accuracy high, while rare classes lower the mIoU.

The loss does not always agree with mIoU. The lowest loss (0.084) belongs to FPN with ResNet-34, which has only 42.9\% mIoU; FPN with EfficientNet-B4 has the second lowest loss (0.092) but the highest mIoU and accuracy. For this reason, mIoU rather than the loss is recommended for choosing the model.

\subsection{Comparison with the BDD100K Reference}
The BDD100K paper reports 56.9\% mIoU for segmentation-only training \cite{yu2020bdd100k}. The best model is 5.7 points higher (62.6\%), and FPN with EfficientNet-B3 is 1.7 points higher. All other 50 models are below this value, by more than 12 points. This comparison is indicative, because the data split, the label set and the evaluation protocol of the benchmark paper may differ in details from those of this study.

\subsection{Scope of the Study}
This work is a controlled empirical comparison: all 52 models use one protocol, so the differences between them come from the choice of backbone and decoder. The main trends are large and clear, for example the gap between FPN with EfficientNet and all other models. Small differences between backbones with similar mIoU (for example 41--43\%) are best read as comparable performance. The evaluation uses mIoU, pixel accuracy and loss; per-class IoU and qualitative examples would add detail on which classes benefit most from a better backbone. The class imbalance of the data can be studied further with class weights or imbalance-aware losses such as the Dice loss \cite{milletari2016vnet} or the focal loss \cite{lin2017focal} (see \cite{jadon2020survey} for a survey). Parameter counts and inference speed are natural additions for in-vehicle use, and transformer-based networks are a natural extension of the same benchmark.

\section{Conclusion and Future Work}
This work compared 52 combinations of four decoders and 13 backbones for semantic segmentation on BDD100K. The choice of backbone has a large effect, and this effect depends on the decoder. FPN with an EfficientNet-B4 backbone was the best model, with 62.6\% mIoU, 89.8\% accuracy and a loss of 0.092, and it was the only model family that exceeded the 56.9\% mIoU of the benchmark reference. Better segmentation gives the vehicle a better understanding of the road scene.

Several directions follow from this work. First, repeat the experiments with several random seeds, add ablations to explain the FPN--EfficientNet interaction, and report per-class IoU, parameter counts and speed. Second, address class imbalance with class weights or other loss functions, and compare with transformer-based networks such as SegFormer \cite{xie2021segformer}. Third, test the best model on adverse conditions: a new snowy-weather dataset could be built, or existing data such as Mapillary Vistas \cite{neuhold2017mapillary} and the Canadian Adverse Driving Conditions dataset \cite{pitropov2021cadc} could be used. Fourth, check whether the model works in multi-task learning and in other domains such as medical images \cite{menze2015brats} and aerial images \cite{zamir2019isaid}.

\bibliographystyle{IEEEtran}
\bibliography{refs}

\end{document}